\documentclass[5p]{elsarticle} % double col
\usepackage{lineno,hyperref}
\usepackage[ruled,vlined]{algorithm2e}
\usepackage{makecell}
\usepackage{multirow}
\usepackage[fleqn]{amsmath}
\usepackage{amssymb}
\usepackage{amsfonts}
\usepackage{enumitem}
\usepackage{pifont}
\usepackage{arydshln}
\usepackage[table,xcdraw]{xcolor}
\usepackage{xcolor}
\usepackage{array}
\graphicspath{{figures/}}
\usepackage[export]{adjustbox}
\usepackage{caption}
\usepackage{subcaption}
\usepackage{tabularray}
\UseTblrLibrary{functional}
\definecolor{Mercury}{rgb}{0.89,0.89,0.89}
\usepackage{threeparttable}
\usepackage{amssymb, bm}
\usepackage{svg}
\usepackage{relsize}
\usepackage[normalem]{ulem}
\usepackage{color}
\usepackage{hhline}
\usepackage{rotating}
\usepackage{colortbl}
\usepackage{booktabs}
\usepackage{siunitx}
\usepackage{booktabs}
\usepackage{xcolor}

\newif\ifshowchanges
\showchangesfalse% set \showchangestrue to show modifs in blue, else \showchangesfalse

\usepackage{color}
\newif\ifdraft
\draftfalse
\drafttrue
\ifdraft

\else

\newcommand{\PF}[1]{}
\newcommand{\JG}[1]{}
\newcommand{\DI}[1]{}
\newcommand{\pf}[1]{}

\fi
\newcommand{\rev}[1]{\ifshowchanges\textcolor{blue}{#1}\else#1\fi}
\journal{Journal of \LaTeX\ Templates}

\biboptions{authoryear}

\begin{document}

\interfootnotelinepenalty=10000
\definecolor{Silver}{rgb}{0.752,0.752,0.752}

\begin{frontmatter}

\title{\rev{Scene-agnostic ALS boresight self-calibration}}

\author[topo]{Aurélien Brun\corref{corauthor}}
\ead{aurelien.brun@epfl.ch}
\author[topo]{Jan~Skaloud}
\ead{jan.skaloud@epfl.ch}
\cortext[corauthor]{Corresponding author}
\address[topo]{EPFL ENAC ESO, Bâtiment GC - Station 18, 1015 Lausanne, Switzerland} 

\begin{abstract}
\rev{ALS boresight calibration has relied for two decades on dedicated flight patterns over structured scenes containing planar surfaces of varied aspect and slope. While reliable, this approach imposes constraints on the scene content and operations, which limits its applicability to boresight recovery within routine mapping missions. We present a practical approach that substantially relaxes these requirements by replacing plane-based constraints with scene-agnostic point-to-point correspondences extracted automatically from overlapping ALS strips. Two complementary formulations are proposed to estimate boresight with laser vector observations: (i) a simpler parametric adjustment utilizing INS/GNSS trajectory; (ii) a rigorous formulation treating GNSS and raw inertial data within an existing factor-graph, i.e. a dynamic network, where boresight is added as an additional parameter. Both formulations are evaluated across four operational ALS flights equipped with five inertial systems, covering a wide range of flight altitudes, overlap geometries, terrain types and inertial sensor classes. The analysis draws a clear boundary between the legacy plane-based conditioning that falls short outside the calibration scenario and the proposed formulations, which either recover or absorb boresight effects under conventional mapping geometry. Among them, the lightweight formulation is sufficient for boresight recovery using tactical and navigation grade inertial sensors, while the general factor-graph approach is clearly superior when the inertial sensor errors are less observable within an optimal smoother. This supports the hypothesis that, for INS/GNSS trajectory of sufficient quality, the boresight calibration can be performed without particular scene prerequisites during routine mapping operations using a minimum of 3–4 overlapping strips, with either proposed formulation. When significant time-varying attitude errors are present in the smoothed INS/GNSS trajectory, the boresight gets recovered correctly only with dynamic network while the simpler approach still has positive, mean attitude error-absorbing effect on the resulting point cloud. The Gauss-Helmert implementation is released as open source, together with the data of three of the four flights.
}

\end{abstract}


\end{frontmatter}
%\linenumbers
\newpage
\section{Introduction}

\subsection{ALS Background}

\noindent\textbf{Sensors.} 
In about 30 years of broader practice, airborne laser scanning (ALS) has defied all odds of extinction driven by the arrival of large-format digital cameras capable of $\approx$2-3 cm GSD\footnote{Ground Sampling Distance} imagery from $\sim1$ km above ground. This is owed to a series of technological evolutions that have, for instance, drastically increased acquisition rates: sending pulses before receiving their echoes (addressing the limiting factor of laser travel-time), employing concurrent laser channels, or using extremely sensitive detector arrays in single photon lidars. These technologies enable point densities ranging from a few tens to hundreds of points per m$^2$ from the same ($\sim1$ km) altitude above ground level (AGL\footnote{Altitude above Ground Level}). While the ALS-derived 3D information is complementary in urban scenery, it remains indispensable over vegetation and the reconstruction of detailed complex 3D terrain. For these reasons, the trend is toward concurrent acquisition of lidar and high-resolution camera data, increasingly integrated into dedicated hybrid sensors such as those recently released by \href{https://www.vexcel-imaging.com/ultracam-dragon/}{Vexcel} or \href{https://leica-geosystems.com/products/airborne-systems}{Leica}. 

\noindent\textbf{Georeferencing.} 
In comparison to the advances in instrumentation, the conceptual framework of ALS point-cloud georeferencing within industrial production systems has evolved relatively little (see \cite{Poppl23Review} for a review). The workflow typically separates an instrument/system \textit{calibration phase}, performed under controlled conditions, from a subsequent \textit{mapping phase} in which only a subset of parameters is re-estimated. The \textit{calibration phase} includes scanner internal orientation, radiometric and system response calibration as well as the mounting parameters between the scanner, inertial measurement unit (IMU) and GNSS antenna(s) - namely the translational (lever-arm) and rotational (boresight) offsets between their respective frames, all assumed rigid. The \textit{mapping} stages are generally (i) IMU/GNSS integration into optimized trajectory, (ii) lidar point cloud creation via direct georeferencing (using optimized trajectory); and possibly a follow-up re-adjustment: (iii-a) the ALS strips or blocks \citep{Kager2004, Ressl2008als, GliraPFG2015}, or (iii-b) portions of trajectory in ALS strips \citep{GliraAdjTimeErr2016} or ultimately (iii-c) re-estimation of complete trajectory considering temporal constraints based on raw-inertial data with lidar-spatial constraints \citep{Zhu21Tightly, brun2022, Poppl24DN}. While the latter can be considered as optimal or rigorous, in terms of an estimation-problem formulation and models on the observed quantities (ranges or laser vectors, raw inertial data, etc.), it leads to a larger number of parameters and poses some practical challenges \citep{CucciEcow2019}. This is one reason why broad industrial adoption has lagged despite the approach being proposed long ago \citep{blazquez2004} with rigorous merits confirmed by simulations \citep{Rouzaud2011}, a gap that partly motivates the present work. %with practical benefits reported also in airborne gravimetry \citep{SkaloudGAL2015} and camera \citep{cucci_rawbundle_2017}. 

\subsection{Practice and limits}\label{sec:practice}

\noindent\textbf{Scan matching}.
The trajectory and point-cloud adjustment methods (iii-a,b,c) differ not only in how they formulate the problem but also in what geometric entities they match and how those matches constrain the optimization. During the dedicated calibration phase, both the scenery and the flight parameters (overlap, altitude, pattern) can be chosen to favor a particular matching strategy. This is not the case in the operational mapping phase, where data is acquired with minimal overlap between successive flight-lines for economic reasons. A matching method suitable for operational use should therefore impose as little configuration prerequisite as possible on the scene content.  

I) Iterative Closest Point (ICP) matching, originally developed for terrestrial laser scanning \citep{ChenICP1991} and suggested also for ALS \cite{GliraRigICP2015} imposes very few prerequisites on the scene context. However, for high-end systems with platform-stabilized scanners, its applicability is limited to residual or relatively \textit{small errors} in the point-cloud, and therefore generally reserved to high quality, well calibrated systems. 

II) Considerably larger errors in the point cloud can be tolerated when parts of the scenery include a sufficient number of regular surfaces, such as planes \citep{Kager2004, Filin2004, Friess2006}. As a planar surface only restrains lidar points in their normal direction, such matching is more suitable for \textit{urban areas} where artificial planar surfaces (e.g. roofs) are present and vary in inclination and orientation. However, some parts of a natural terrain may be approximated by so-called ``feature planes'', possibly triangular, to which a point (or group of points) are matched \citep{Kerstling2012, Jonassen23Vox, Poppl24DN}. Such an approach is less restrictive on the scene topology yet \textit{introduces approximation errors inherent to the planar fit} and still constrains only in the directions of surface normals. Consequently, the terrain has to \textit{vary significantly in steepness and orientation} to fully constrain the adjustment. 

III) Albeit geometrically stronger as constraints, the proposed linear-feature matching techniques \citep{Lie2007LinFea} have received less attention in practice, probably due to its \textit{limitation to urban scenarios} where tilted planes are also abundant.

IV) The application of deep learning for robotic SLAM, terrestrial kinematic scanning and automated driving, offered new techniques in detecting correspondences across optical sensor data, which we adapted for detecting directly point-to-point correspondences within overlapping portion of ALS strips \citep{brun2022}. This approach employs learned descriptors that express information about structural characteristics within a point's neighborhood in latent space. Similar characteristics of such ``3D keypoints'' are matched among overlapping portions of the scan and can in principle be identified across large georeferencing errors (in point clouds generated by non-calibrated systems or with significant trajectory errors, see Sec.~\ref{sec:res}). Akin to ICP, this \textit{constraint acts in all directions} and is therefore considerably stronger than methods relying on planar surfaces. Unlike ICP, it tolerates much larger point-cloud errors. It is also robust to varying point density on surfaces with low reflection or structural information. It has also been demonstrated that such ``3D keypoints'' can be effectively applied in the rigorous re-optimization of the full acquisition trajectory (iii-c) even in the case of irregular spacing.
%Its irregular spacing is, however, generally not limiting factor when used in the rigorous re-optimization of the whole trajectory (i.e. iii-c like). 
%It detection and application was recently improved, released as an open-source, and its generality tested on kinematic scanners ranging from handheld to airborne platforms \citep{Brun2025gen}. 
Recently improved in the detection and application, the method reaches accuracy of half the point-sampling density or better. Tested and validated for general application on kinematic scanner data ranging from handheld to airborne platforms, the method was recently released as open-source \citep{Brun2025gen}. Among the reviewed matching strategies, this method imposes the least prerequisite on scene content, making it suitable for general operational use.

\vspace{1mm}
\noindent\textbf{Navigation sensors.} %Unless artificially perturbed, the quality of signal reception from satellite positioning is usually less problematic in ALS, and with the various global or wide-area differential correction services, it is ``just the quality'' of IMU, respectively, its attitude that has to meet the fantastic ranging accuracy of modern lidar: $2-5$ cm ranging accuracy with only $0.1-0.2$ footprint at $1$ km (AGL\footnote{Altitude above Ground Level}). Yet $0.05$ m error over $1$ km corresponds to $0.05$ mrad ($\approx 0.003$ deg) pointing precision, which is the orientation noise level of the ``well aligned'' navigation-grade IMU in roll and pitch; the very best IMU that civilian users may get their hands on, albeit not without a special authorization or operational restrictions. However, the heading uncertainty for those instruments are (at the best about) 2 times lower than in roll and pitch on what is considered ``the typical mission profile''\footnote{Citation from  \hyperlink{https://applanix.trimble.com/en/products/hardware/applanix-pos-av}{Trimble Applanix POS AV 610} product.}, yet practically often more on longer flight-lines. To avoid the annoyance with ALS operational restrictions hosting a navigation grade IMU, some high-end lidar systems employ IMUs with gyro drifts $>0.01$ deg/h which is in turn re-defying what such ``typical mission profile'' should be to limit its influence on lidar point cloud. 
Accurate airborne navigation relies on two primary inputs: satellite signal positioning and IMU measurements. Unless artificially perturbed, obtaining satellite positioning signals is rarely the problem once the platform is above the terrain, and thanks to the widespread availability of precise wide-area satellite services \citep{WideRTK2010} it can be obtained globally with high-quality. The real challenge, and primary technical bottleneck, is therefore the IMU data: to match the high precision of modern lidars (2-5~cm ranging accuracy with only 0.1-0.2~m laser-footprint at $1$ km AGL), the inertial sensors must be of exceptionally high-quality. Indeed $0.05$ m error seen from $1$ km corresponds to $0.05$ mrad ($\sim0.003^\circ$). This is at the \textit{limit of a pointing precision} (orientation noise level) of a ``well aligned'' navigation-grade IMU, the very best IMU that civilian users may access, albeit not without special authorization nor operational restrictions. This noise level is, however, only achievable in roll and pitch angles. The yaw (heading angle) of such instruments is at best twice as large under a ``typical mission profile''\footnote{Citation from \href{https://applanix.trimble.com/en/products/hardware/applanix-pos-av}{Trimble Applanix POS AV 610} product.} - and \textit{practically worse} on longer flight-lines where observability of inertial errors degrades. 
To avoid the annoyance of operational restrictions when hosting a navigation grade IMU, we should also note that some high-end mapping systems employ IMUs with gyro drifts $>0.01^\circ/$h, which in turn substantially shortens the flight-line duration of a ``typical mission profile'' to limit the heading error to tolerable levels. 
%; ii) a compromise is made for a smaller, lighter and thus less precise IMU to accommodate some other constraints, e.g. instrument structural safety, payload volume, weight or cost. 

The lower performance in heading stems from the naturally limited observability of errors in this direction within an IMU/GNSS system, which further deteriorates when the IMU is placed on a (now widely used) stabilized platform. Indeed, platform stabilization reduces the rotational dynamics that the system relies on to separate and dampen attitude errors. These errors, arising from both accumulated gyroscope noise and imperfect initial alignment can only be corrected gradually through IMU/GNSS integration, requiring at least 15-20 min of flight to converge. 
%The reason behind this is the naturally limited observability of heading errors within an INS/GNSS system due its imperfect initialization that is further deteriorated when the IMU is placed on (now widely used) stabilized platform. 
More specifically, the heading errors in such systems can only be observed during a flight when manifested as velocity errors and de-correlated with those caused by accelerometers. This necessitates that measurements of \textit{all} accelerometers are non-zero and that rotational dynamics are present \citep{Hong2025ObsAnal}. On a flight-line however, an IMU placed on a stabilized platform flown at constant velocity and direction is experiencing forces that are practically indistinguishable from static conditions. %\footnote{This was illustrated in practice when An earlier firmware of our modern INS erroneously identified conditions for a zero velocity update while being stabilized airborne!}.
In summary, current \textit{navigation technology is limited in attitude precision} for ALS even when equipped with a high-end, navigation-grade IMU while the use of platform stabilization decreases the ability to reduce attitude errors via GNSS data.
% on a stabilized airborne!} 
%This fact has an impact on boresight calibration due to its correlation with residual attitude error caused either by its imperfect initialization (so called initial alignment) or gyro-drift. Note, that the former is constantly re-estimated by the INS/GNSS integration and requires at least 15 min to be dampened by GNSS data in flight, or for a high-end IMU by stationary observations prior to flight. It is therefore not the best (yet still existing) practice to power up an ALS system comprising an IMU just few minutes before the first flight-line (or switching it off right after the last flight-line). 

\vspace{1mm}
\noindent\textbf{Boresight.} 
The boresight estimation method depends fundamentally on what is assumed about the trajectory. It can be estimated (1) by considering a perfect trajectory, i.e. sufficiently precise, (2) together with trajectory-correcting parameters (iii-a,b), or (3) via the re-estimation of the whole trajectory from GNSS and raw IMU data through rigorous modeling of all causes (iii-c). In all cases, incorrect or zero system boresight knowledge has a very large impact on the accuracy and consistency of the georeferenced point-cloud (e.g. $\sim 50$ m for $3^\circ$ at $1$ km AGL). For these reasons the boresight calibration is initially executed in a scenario where the positioning errors in the trajectory are negligible and the remaining attitude bias is small. 

While large discrepancies in point clouds across flight-lines are less suitable for ICP or point-to-patch constraints, matching between planar surfaces differing in size and orientation is more robust \citep{Skaloud2006b} and can be automatized \citep{Tarsha2007HoughT}. For the same reason it is used by industry for calibrating not only boresight but possibly also some parameters in scanner geometry \citep{RieglMan19}. To decouple the boresight angles and minimize the impact of residual attitude errors, the boresight calibration is initially executed in a cross-pattern on flight-lines of shorter duration flown at 2-3 flight levels over an urban scene \citep{Skaloud2007a, RieglMan19}. Albeit demanding in practical execution, this approach is generally reliable and has evolved little conceptually during the last 20 years. However, it is important to note that this approach does not address the possibly significant value of attitude bias (e.g. due to imperfect IMU initialization), the mean value of which gets absorbed by the estimated boresight. This is, among others, the limiting factor when calibrating the boresight in the laboratory as suggested by \cite{Baumker2001}.   

\subsection{Contribution}
\rev{We present a practical workflow for ALS boresight self-calibration that substantially relaxes the scene requirements of conventional plane-based approaches and demonstrate its practical applicability within mapping operations under nominal reception of GNSS signals.  
Point-to-point correspondences between overlapping strips are extracted automatically using the LiMatch method \citep{Brun2025gen}. They are then used as geometric constraints in two complementary formulations: 
The first is a Gauss-Helmert (GH) adjustment that treats the INS/GNSS trajectory, determined via an optimal smoother, as fixed observations and estimates the three boresight angles as the only unknowns. This approach is simpler, executes rapidly, and is thus accessible to any operator with a post-processed INS/GNSS trajectory. As shown later it can be used, depending on the attitude quality, either for boresight calibration or for its absorption together with a mean attitude error.  
\\
The second approach leverages the more involved formulation via the Dynamic Network (DN) framework with point-to-point constraint as presented in \cite{Brun2025gen}, in which the boresight angles are formulated as additional parameters. This path provides a joint rigorous modeling and estimation of trajectory, boresight and time-correlated inertial errors as parameters. Both paths share the same scene-agnostic point-to-point correspondences as input, and are evaluated together across five system configurations on four operational ALS flights and compared to conventional plane-based approaches. In summary, the contributions of this paper are:
\begin{itemize}
    \item A simpler parametric, Gauss-Helmert (GH) formulation of boresight calibration using point-to-point correspondences that removes the former requirement for a scenery rich of planar surfaces \citep{Skaloud2006b},
    \item The first validation of boresight estimation within the rigorous modeling of Dynamic Network across the defined operational minimum on various ALS configurations ranging from UAV, through tactical, to navigation grade systems, including a per-configuration analysis of boresight-to-inertial-bias decorrelation,
    \item The definition of a minimal requirement on composition of flight-lines for separating boresight angles from time correlated inertial sensors without dedicated calibration patterns, 
    \item The validation of boresight estimation with the simpler GH formulation over the same operational envelope for navigation and tactical grade inertial systems,    
    \item A demonstration of the capacity of the GH formulation to absorb the boresight together with the mean attitude error on a UAV-grade system, yielding a usable point cloud whose residuals remain roughly twice those of its navigation-grade counterpart over identical geometry,
    \item Openly released data from three experiments on all levels: raw inertial, GNSS and ALS, point correspondences, approximate and adjusted trajectories and point clouds,
    \item An open-source implementation of the Gauss-Helmert recovery/absorption of boresight with experiment related configurations.
\end{itemize}
}

\subsection{Paper organization.} 
Sec.~\ref{sec:model} reviews the approach of establishing the point-to-point correspondences in the overlapping part of the strips, describes the lightweight Gauss-Helmert functional model based on an optimal INS/GNSS trajectory\rev{, as well as the rigorous formulation via Dynamic Network utilizing GNSS and raw inertial data.} Sec.~\ref{sec:exp} introduces the four flights and five system configurations, spanning the legacy calibration configuration from \cite{Skaloud2006b}, corridor mapping and conventional block mapping over peri-urban and natural terrain with Riegl and Leica systems. Sec.~\ref{sec:res} presents the results in terms of the stability of model convergence and goodness of the recovered parameters. Sec.~\ref{sec:disc} discusses the consistency and respective operating domains of the two formulations and compares them against the traditional plane-based approach.

\section{Adjustment model}\label{sec:model}

\subsection{Point-to-point correspondences}\label{sec:p2p}
The proposed adjustment relies on automatically established point-to-point correspondences between overlapping ALS strips. These correspondences provide the geometric constraints required to estimate the boresight angles within a least-squares framework. A detailed description of the extraction procedure is provided in \citet{Brun2025gen}; only the essential concepts are summarized here.

The objective is to identify pairs of \textit{key points} acquired in different flight lines that correspond to the same physical features, independently of residual georeferencing errors. In contrast to traditional plane-based calibration approaches, no assumption is made regarding the presence of planar or otherwise parametrizable surfaces. The constraints are therefore identified directly from discrete points (and their neighborhood) within the overlapping clouds.

Conceptually, the procedure parallels tie-point extraction in photogrammetry but operates on point clouds rather than images. It comprises automated keypoint detection, local geometric description, descriptor-based matching across strips, robust outlier rejection, and subsequent local geometric refinement. Each correspondence introduces a three-dimensional consistency constraint directly between \textit{two} discrete lidar measurements and is therefore not affected by surface approximation. Once established, these correspondences constitute the \rev{geometrical constraints that can be introduced as observation equations in the Gauss-Helmert model (Sec.~\ref{sec:funcmodel}), or within a Dynamic Network (Sec.~\ref{sec:dn_model}).}

\subsection{Gauss-Helmert functional model}\label{sec:funcmodel}
The functional model enforces geometric consistency between corresponding lidar points by minimizing the differences between their coordinates in the mapping frame after direct georeferencing. Let $m$ denote an arbitrary Cartesian mapping frame. The direct georeferencing of a lidar measurement acquired at epoch $k$ using a navigation system (INS/GNSS) mounted on the same platform can be expressed as:
\noindent
\begin{equation}\label{eq:dgf}
    \mathbf{p}_k^m = \mathbf{n}_k^m+\mathbf{R}_{b,k}^m(\bm{\omega}_k)\,
    \left[ %\begin{pmatrix}
        \begin{pmatrix}
            \mathbf{I} + \bm{\Omega}_{b^*}^b
        \end{pmatrix}
        \mathbf{T}_s^{b*}\mathbf{u}^s_k+\mathbf{a}^b
  \right]  %\end{pmatrix}.
\end{equation}
Under the small-angle assumption, neglecting 2nd order and higher terms allows us to linearize the boresight rotation:
\begin{equation}\label{eq:dgf2}
    \mathbf{p}_k^m = \mathbf{n}_k^m+\mathbf{R}_{b,k}^m(\bm{\omega}_k)\,\Big(\mathbf{u}_k^b+ [\bm\theta]_{\times}\mathbf{u}^b_k +\mathbf{a}^b\Big).
\end{equation}
The variables in Eq.~(\ref{eq:dgf}) and (\ref{eq:dgf2}) are defined as follows:

\noindent
\begin{tabular}{@{}l p{0.7\linewidth}@{}}
    \toprule
    $\mathbf{p}_k^m$ & Lidar point coordinates in the mapping frame at epoch $k$ \\
    $\mathbf{n}_k^m$ & Platform position in mapping frame (INS/GNSS) at epoch $k$\\
    $\mathbf{R}_{b,k}^m(\bm{\omega}_k)$ & Body-to-map rotation at epoch $k$ \\
    $\bm{\omega}_k = [r_k\, p_k\, y_k]$ & Roll, pitch, yaw angles at epoch $k$ \\
    $\bm\theta = [\alpha\,\beta\,\gamma]^T$ & Boresight small misalignment angles \\
    $\bm{\Omega}_{b^*}^b$ & Lidar boresight matrix \\
    $[\bm\theta]_{\times}$ & Skew-symmetric matrix part of $\bm{\Omega}_{b^*}^b$ \\
    $\mathbf{T}_s^{b*}$ & A priori known rotation from lidar to body frame \\
    $\mathbf{u}_k^s$ & Lidar measurement in sensor frame \\
    $\mathbf{u}_k^b$ & Lidar measurement in a priori known body frame \\
    $\mathbf{a}^b$ & The spatial offset (so-called lever-arm) in body frame \\
    \bottomrule
\end{tabular}
\newline

\noindent
To express Eq.~\ref{eq:dgf2} as a linear function of the boresight angles, the small angle approximation and the anti-commutative property of the cross-product are used:
\begin{equation}
    [\bm\theta]_{\times}\mathbf{u}^b_k = -[\bm u^b_k]_{\times}\bm{\theta} \doteq \mathbf{U}^b_k\bm{\theta} = \begin{bmatrix}0 & w_k & -v_k \\ -w_k & 0 & u_k \\ v_k & -u_k & 0\end{bmatrix}\bm{\theta}.
\end{equation}
Substituting this relation into Eq.~\ref{eq:dgf2} yields the linearized georeferencing equation
\begin{equation}\label{eq:lgf}
    \mathbf{p}_k^m 
    = \mathbf{n}_k^m 
    + \mathbf{R}_{b,k}^m(\bm{\omega}_k)\,\Big(\mathbf{u}^b_k + \mathbf{U}^b_k\bm{\theta} + \mathbf{a}^b\Big).
\end{equation}

\begin{figure}[h]
    \centering
    \includegraphics[]{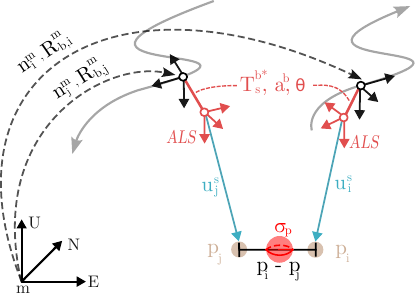}
    \caption{Graphical representation of a point-to-point constraint.}\label{fig:p2p}
\end{figure}

\noindent The objective is to estimate the boresight angles $\bm{\theta}$ such that the coordinate discrepancies between corresponding points in overlapping strips are minimized, as illustrated in Fig.~\ref{fig:p2p}.

\noindent For a correspondence between two measurements acquired at epochs $i$ and $j$, the following zero-valued condition equation is formulated:
\begin{equation}\label{eq:f}
    0 = f(\bm{\theta}, \bm{\ell}) + \mathbf{w} = \mathbf{p}_i^m - \mathbf{p}_j^m + \mathbf{w}.
\end{equation}
\noindent Substituting Eq.~\ref{eq:lgf} into Eq.~\ref{eq:f} shows that the coordinate difference depends on the boresight parameters $\bm{\theta}$ and on the navigation observations provided by the INS/GNSS, $\bm{\ell}_{ij}$ at epochs $i$ and $j$, defined as
\small{
\begin{equation}\label{eq:l}
    \bm{\ell}_{ij} =
    \big[X_i^m,\,Y_i^m,\,Z_i^m,\,r_i,\,p_i,\,y_i,\,
          X_j^m,\,Y_j^m,\,Z_j^m,\,r_j,\,p_j,\,y_j\big]^T
\end{equation}
}

\noindent For a set of $p$ point-to-point correspondences, the adjustment problem comprises
$m = 3p$ condition equations (3 conditions per correspondence, one per coordinate),
$u = 3$ unknown parameters,
$n = 12p$ observations, 
and no explicit constraints ($c = 0$). 
The resulting degrees of freedom are therefore
\[
r = m - u + c = 3(p-1).
\]
As a result, the redundancy increases linearly with the number of correspondences, contributing to the statistical robustness of the adjustment. In practice however, the estimation remains stable even with a strongly reduced correspondence set, provided that sufficient geometric diversity is preserved in the matched points. This property is demonstrated by the frequency downsampling experiment in Sec.~\ref{sec:down_freq}, where accurate estimates are obtained even for very sparse subsets of correspondence.

\subsection{Gauss-Helmert linearized model}\label{sec:linmodel}
Since both deterministic boresight parameters and stochastic navigation observations enter the condition equations, the problem is formulated within a combined Gauss-Helmert model. Linearization of Eq.~\ref{eq:f} around initial estimates yields
\begin{equation}\label{eq:f_lin}
    \mathbf{A}\,\Delta\bm{\theta}
    + \mathbf{B}\,\mathbf{v}
    + \mathbf{w}
    = \mathbf{0},
\end{equation}
where $\mathbf A = \partial f / \partial \bm{\theta}$ and $\mathbf B = \partial f / \partial \bm{\ell}$ are the Jacobian matrices with respect to the parameters and observations, respectively; $\Delta\bm{\theta}$ denotes the parameter corrections, $\mathbf{v}$ the observation corrections, and $\mathbf{w} = f(\hat{\bm{\theta}},\hat{\bm{\ell}})$ the misclosure vector. Explicit expressions of the Jacobians are provided in the Appendix. 

\noindent\textbf{Stochastic model.}
In the classical Gauss-Helmert formulation, stochasticity is introduced through the observations, while the condition equations themselves are treated as deterministic. In the present work, however, the point-to-point consistency equations are also assigned stochastic weights to account for residuals arising from the scanning and correspondence extraction process \citep{Brun2025gen}. This treatment is conceptually similar to the weighting of tie-point constraints in photogrammetric bundle adjustment, where automatically extracted image correspondences are also affected by measurement and matching uncertainties.

It is known that the navigation solution provided by the INS/GNSS integration exhibits temporal correlations \citep{Skaloud2003a}. For computational efficiency, these correlations are neglected and their effect is absorbed either through appropriately inflated variances or through correspondences selection (Sec.~\ref{sec:down_freq}). This approximation is further justified because correspondences are typically established between spatially separated epochs belonging to different flight lines, thereby reducing the influence of short-term temporal correlations. All observation errors are assumed zero-mean and mutually uncorrelated. As a result, the weight matrices of the observations $\mathbf{P}$ and of the conditions $\mathbf{W}$ are diagonal:

\begin{equation}
    \mathbf{P} = 
    \begin{bmatrix}
        \mathbf P_1 & \ldots & 0 \\
        \vdots  & \ddots &\vdots \\
        0  & \ldots & \mathbf P_p
    \end{bmatrix}
    \text{, }
    \mathbf{W}=
    \begin{bmatrix}
        \mathbf W_1  & \ldots & 0 \\
        \vdots  & \ddots &\vdots \\
        0  & \ldots & \mathbf W_p
    \end{bmatrix}
\end{equation}
where each block $\mathbf P_i$ contains the inverse variances of the twelve navigation quantities associated with correspondence $i$ while $\mathbf W_i = \sigma_p^{-2}\mathbf{I}_3$ reflects the equal weighting of the three coordinate components of each point-to-point constraint, where $\sigma_p$ represents the standard deviation assigned to the point-to-point consistency constraint. The weight therefore reflects the expected uncertainty of the geometric correspondence between the two lidar points.

\noindent\textbf{Objective function.} The adjustment minimizes the quadratic objective function
\begin{equation}
J(\mathbf{v},\Delta\bm{\theta}) =
\mathbf{v}^\top \mathbf{P}\mathbf{v}
+
\big(\mathbf A\Delta\bm{\theta} + \mathbf B\mathbf{v} + \mathbf{w}\big)^\top
\mathbf{W}
\big(\mathbf A\Delta\bm{\theta} + \mathbf B\mathbf{v} + \mathbf{w}\big).
\end{equation}
The system therefore consists of $3p$ condition equations linking the navigation observations and the boresight parameters through the point-to-point constraint.
Setting the first derivatives with respect to $\mathbf{v}$ and $\Delta\bm{\theta}$ equal to zero yields the normal equations of the Gauss-Helmert model (see, e.g., \citealp[Ch.~5]{Teunissen2000}). 
Eliminating the observation corrections $\mathbf{v}$ leads to a reduced system in $\Delta\bm{\theta}$, which is solved iteratively until convergence.

After convergence, the estimated boresight parameters are obtained as
$
\hat{\bm{\theta}} = \bm{\theta}_0 + \Delta\bm{\theta}.
$
Because the initial approximation is set to $\bm{\theta}_0 = \mathbf{0}$, the estimate reduces to $\hat{\bm{\theta}} = \Delta\bm{\theta}$. Finally, the posterior covariance matrix of the parameters is obtained from the inverse of the reduced normal matrix.

\subsection{Rigorous formulation via Dynamic Network}\label{sec:dn_model}
\rev{
The point-to-point correspondences of Sec.~\ref{sec:p2p} can equally be introduced as geometric constraints in a rigorous trajectory-estimation framework, where boresight is recovered jointly with the trajectory rather than from a trajectory assumed fixed. We rely on the Dynamic Network (DN) formulation of our previous work \citep{brun2022, Mouzakidou24, Brun2025gen} summarized here with one key difference: the lidar boresight is kept as a free parameter, here recovered from the point-to-point constraints alone and reported rigorously, whereas in \citet{brun2022} it was estimated without rigorous reporting of the angles and in \citet{Mouzakidou24} only jointly with camera tie-points, in both cases in the same single flight. %For the full dynamic network (general form of factor-graph) machinery we refer the reader to those publications.
}

\noindent \rev{
A Dynamic Network is a general factor graph \citep{loeliger2007factorgraph} representing the joint probability of the sensor measurements, some of them being derivatives of estimated parameters, given the unknown trajectory, calibration and mounting parameters. Its nodes are the body-frame poses at discrete epochs $\Gamma^m_{b,k}=\left[ \mathbf{n}_k^m,\, \mathbf{R}_{b,k}^m \right]$, the time-dependent inertial sensor errors, and in this case also boresight $\mathbf{\theta}$ as the system calibration parameter; its edges are the sensor measurement models, comprising the raw inertial observations, the GNSS positions and/or velocities, and the lidar point-to-point correspondences. Maximizing the joint likelihood yields an optimal estimate of all unknowns simultaneously. The raw inertial and GNSS edges and the associated inertial error models are formulated as in \cite{cucci_rawbundle_2017} %\citep{Mouzakidou24, Brun2025gen} 
and are not reproduced here.
}

\noindent \rev{The point-to-point measurement model is the only edge specific to the present problem. In analogy to Eq.~\ref{eq:f}, given a correspondence between two points $\bm p^m_i, \bm p^m_j$ emitted at epochs $i, j$ with lidar vectors $\bm u^s_i, \bm u^s_j$ in the lidar frame, the constraint depicted in Fig.~\ref{fig:p2p} requires that the two vectors coincide once expressed in the mapping frame:
\begin{equation}
    \Gamma^m_{b,i}\, \Gamma_s^b\, \bm u^s_i - \Gamma^m_{b,j}\, \Gamma_s^b\, \bm u^s_j = \bm \xi,\qquad \bm \xi \sim \mathcal{N}(0, \sigma_p^2 \mathbf{I}_{3\times3}),
    \label{eq:dn_p2p}
\end{equation}
where $\Gamma^m_{b,k} \in \mathrm{SE}(3)$ is the body-frame pose in the mapping frame at epoch $k$, $\Gamma_s^b=\left[ \mathbf{a}^b,\, [\theta]_{\times} \mathbf{T}_s^{b*}\right]$ the sensor-to-body transformation containing the lever-arm and boresight, and $\bm \xi$ a zero-mean Gaussian noise of standard deviation $\sigma_p$, assigned as in the GH formulation from the correspondence quality (Sec.~\ref{sec:linmodel}). The boresight enters Eq.~\ref{eq:dn_p2p} through $\Gamma_s^b$; setting it as a free parameter adds a single boresight node to which every lidar edge connects, leaving the rest of the graph unchanged. The two formulations therefore share the same point-to-point observation and the same boresight unknown, yet differ substantially in what else is estimated: the GH adjustment of Sec.~\ref{sec:funcmodel} treats the smoothed INS/GNSS trajectory as fixed observations with known noise and solves for the boresight alone, whereas the DN estimates the boresight jointly with the trajectory and the time-correlated inertial errors using raw data from gyroscopes and accelerometers. This makes DN a broader adjustment with rigorous modeling of the IMU observations, the influence of which varies according to inertial sensor quality as examined in Sec.~\ref{sec:res}.}
\section{Experiments}\label{sec:exp} 
\subsection{Datasets and geometric configuration}
\begin{figure}[ht!] 
\centering 
\includegraphics[width=0.94\linewidth]{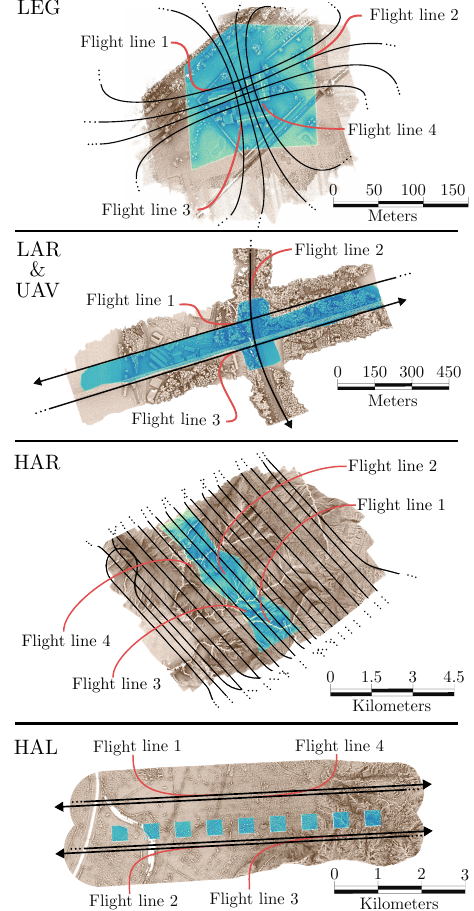}
\caption{Geometries of the four studied flights. The LAR flight carries both the navigation-grade (LAR) and UAV-grade (UAV) configurations. Full point clouds and overlapping sections are depicted in brown and blue, respectively. Trajectories are shown in black.\label{fig:areas}}
\end{figure}

The robustness and observability of the proposed calibration approach are evaluated on four operational ALS flights carrying five inertial system configurations representing a broad range of geometric configurations, sensor classes and scenery conditions. The main sensor and flight characteristics are summarized in Table~\ref{tab:flight_param}, while flight geometries and overlap regions used for constraint extraction are shown in Fig.~\ref{fig:areas}.

\begin{table*}[t]
\centering
\footnotesize{
\begin{tblr}{
  width = 0.95\linewidth,
  colspec = {Q[20]Q[180]Q[115]Q[125]Q[125]Q[125]Q[125]},
  cells = {c},
  cell{2}{1} = {r=2}{},
  cell{4}{1} = {r=3}{},
  cell{7}{1} = {r=7}{},
  cell{2}{4} = {c=2}{},
  cell{3}{4} = {c=2}{},
  cell{7}{4}  = {c=2}{},
  cell{8}{4}  = {c=2}{},
  cell{9}{4}  = {c=2}{},
  cell{10}{4} = {c=2}{},
  cell{11}{4} = {c=2}{},
  cell{12}{4} = {c=2}{},
  cell{13}{4} = {c=2}{},
  hline{2,4,7,14} = {-}{},
  hline{3,5-6,8-13} = {2-7}{Silver},
}
 &  & {\textbf{Legacy}\\ \textbf{(LEG)}}
   & {\textbf{Low altitude}\\ \textbf{Riegl (LAR)}}
   & {\textbf{Low altitude}\\ \rev{\textbf{UAV (UAV)}}}
   & {\textbf{High altitude}\\ \textbf{Riegl (HAR)}}
   & {\textbf{High altitude}\\ \textbf{Leica (HAL)}}\\
\begin{sideways}\textbf{Flight}\end{sideways}
 & Environment  & built-up  & mixed     &  & natural     & mixed   \\
 & AGL~[m]      & 100--200  & 150--250  &  & 1000--2000  & 2000+   \\
\begin{sideways}\textbf{Navig.}\end{sideways}
 & IMU~grade (Type)  & Tactical (LN200) & Navigation (AIRINS) & UAV (APX-15) & Navigation (IMU57) & High~tactical (ISA-100C) \\
 & Pos.~RMSE~[m]     & \textless{}5 cm   & \textless{}5 cm      & \textless{}5 cm         & \textless{}5 cm     & \textless{}5 cm           \\
 & rp,y~RMSE~[°]     & 0.005,~0.010     & 0.002,~0.005        & 0.025,~0.080           & 0.0025,~0.005      & 0.005,~0.008             \\
\begin{sideways}\textbf{Lidar}\end{sideways}
 & Model                           & LMS-Q240i & VQ480 &  & VQ-1560 II      & TerrainMapper-3 \\
 & Used PRR~[kHz]                  & 30        & 200   &  & 1$\times$1000  & 2000            \\
 & Mean GSD~[m]                    & 0.50      & 0.20  &  & 0.40           & 0.25            \\
 & FOV~[°]                         & 60        & 75    &  & 58             & 60              \\
 & Across-track swath~[m]          & 150       & 250   &  & 1500           & 2500            \\
 & Beam div.~[mrad]                & 2.7       & 0.35  &  & 0.18           & 0.12            \\
 & Mean point ground footprint~[m] & 0.40      & 0.04  &  & 0.35           & 0.25
\end{tblr}}
\caption{Flight and sensor characteristics of the studied setups. The UAV configuration
reuses the LAR flight geometry and scanner with a UAV-grade APX-15 IMU in place of
the navigation-grade AIRINS.}
\label{tab:flight_param}
\end{table*}

The four flights span a wide spectrum of ALS operations and technology. They include both legacy and modern lidar systems, ranging from the 30\,kHz LMS-Q240i to > 2\,MHz in VQ-1560 II and TerrainMapper-3, and cover flight altitudes from 100\,m to more than 2000\,m above ground. The surveyed scenarios vary from structured urban areas to highly mountainous terrain. These variations result in substantial differences in ground sampling distance (GSD), swath width, beam footprint and surface structure, which in turn lead to significantly different conditions for boresight estimation.

\noindent The four flights also differ substantially in flight configuration: the LEG dataset consists of eight short flight lines acquired in two low altitude blocks over a built-up area. The resulting overlap forms a compact square region of 250\,m $\times$ 250\,m. The LAR dataset includes two parallel strips and one cross strip flown at a different altitude, producing strong intersection geometry in otherwise moderately flat terrain. %\rev{Additionally, the LAR
\rev{In addition to the navigation-grade AIRINS, the LAR flight carried a UAV-grade APX-15 IMU (Applanix), sharing the same GNSS and the same lidar. This second IMU configuration, denoted UAV, demonstrates the adequacy of calibration via DN when non-negligible time-correlated errors remain in the trajectory while characterizing the GH behavior in such conditions.} HAR comprises twenty-one parallel strips acquired over highly mountainous terrain with elevation changes higher than 1\,km. Finally, HAL consists of four parallel strips flown at constant altitude over mixed structured urban and natural areas.

\subsection{Processing methodology} \label{sec:processing}
\noindent\textbf{Reference boresight.}
For the LEG, LAR and UAV datasets, reference boresight values are obtained using a conventional point-to-plane calibration approach \citep{Skaloud2006b} applied in dedicated calibration flights. These calibration flights consist of two cloverleaf patterns acquired at different altitudes over structured urban areas containing well-defined tilted planar surfaces. For HAR and HAL, the angles are provided by the manufacturer's own calibration, for the former also based on point-to-plane. These values are considered the most reliable standard for boresight estimation, and thus serve as the reference for assessing the performance of the investigated approaches. \rev{It should be noted, however, that this reference is itself subject to observability limitations, in particular: yaw estimation may be affected by flight altitude and the geometric configuration within and between the calibration strips, and should therefore be interpreted as a high-quality practical standard rather than an absolute ground truth.}

\medskip\noindent\textbf{Direct georeferencing and correspondence extraction.}
For the proposed approach, each dataset is first georeferenced directly using the INS/GNSS smoothed trajectory while assuming zero boresight (only the a priori known cardinal rotations of laser mounting are applied). Depending on the scenario, overlapping regions between two to four successive strips are then identified and point-to-point correspondences are automatically extracted following the procedure described in Sec.~\ref{sec:p2p} and detailed in \cite{Brun2025gen} with the algorithm parameters selected according to the guidelines therein. These are mainly based on scanning density and the approximate object scale. Hence the numerical values differ across datasets but the parameter selection remains consistent, following the same rule-of-thumb criteria.

\medskip\noindent\rev{\textbf{Lightweight calibration.} From the identified correspondences, a random subset of 5000 pairs is selected. This limits the computational load while preserving sufficient redundancy of observations introduced into the Gauss-Helmert adjustment model described in Sec.~\ref{sec:model}.} The stochastic modeling follows the formulation introduced in Sec.~\ref{sec:linmodel}. The observation weight matrix $\mathbf P$ is constructed from the nominal position and attitude variances predicted for each INS/GNSS system, 
%by the manufacturer based on the IMU noise levels (see Tab.~\ref{tab:flight_param}), 
assuming temporally uncorrelated noise. 
The condition weight matrix $\mathbf W$ represents the uncertainty associated with the point-to-point constraints. Following the empirical analysis reported in \cite{Brun2025gen}, the standard deviation of a point-to-point correspondence is assumed to be proportional to the ground sampling distance, $\sigma_{p2p} = \frac{1}{2}\mathrm{GSD}$. Assuming isotropic coordinates in the three coordinate components, the per-coordinate standard deviation becomes $\sigma_p = \sqrt{\frac{1}{3}}\sigma_{p2p} = \frac{1}{2}\sqrt{\frac{1}{3}}\mathrm{GSD}$.

\medskip\noindent\textbf{Rigorous calibration.}
\rev{
In this formulation, the boresight is estimated jointly with the trajectory (Sec.~\ref{sec:dn_model}), from raw inertial and GNSS data rather than from the smoothed trajectory. The same point-to-point correspondences used in the GH adjustment are introduced as lidar edges; the boresight is set as a single free node shared by all lidar edges, while the trajectory poses and the time-dependent inertial errors are estimated simultaneously when solving the large Dynamic Network. The only configuration change relative to \cite{Brun2025gen} is the introduction of the boresight as a free parameter in the factor-graph. The point-to-point noise $\sigma_p$ is set identically to the GH case, so that the two formulations differ only in their treatment of the trajectory and not in the weighting of the shared lidar observations. The boresight posterior covariance recovered from the factor-graph provides the a posteriori correlations among the boresight angles and between the boresight angles and the inertial sensor biases. These are analyzed in Sec.~\ref{sec:res}.}

\medskip\noindent\textbf{Baseline plane-based on-the-job calibration.}
Where the data permits (LEG, LAR, \rev{UAV} and HAR), a baseline plane-to-plane, on-the-job calibration is additionally performed using the manufacturer's software (RiPROCESS, RIEGL). This allows the proposed point-to-point formulations to be compared relative to a conventional plane-based adjustment under identical configurations.

\subsection{Experimental scenarios}
The experimental scenarios serve to assess the estimation accuracy, parameter observability, and convergence behavior of the two proposed boresight calibration paths. The first scenario evaluates the on-the-job calibration itself, applying both the Gauss-Helmert and the Dynamic Network formulations across all datasets; the remaining three characterize the operational robustness of the lightweight Gauss-Helmert path specifically.

\medskip\noindent\textbf{On-the-job calibration.}
In the first scenario, boresight angles are estimated under nominal operations from directly georeferenced strips obtained with zero boresight. The datasets are grouped by inertial-sensor grade, from navigation through tactical to UAV grade, and for each the boresight is recovered by both the GH and the DN formulations from the same point-to-point correspondences. To investigate the influence of overlap geometry, two configurations are considered for each dataset:
\begin{itemize}[noitemsep,topsep=0pt]
    \item \emph{Reduced configuration}, comprising the minimal number of overlapping strips (two flight lines)
    \item \emph{``Sufficient'' configuration}, incorporating additional strips (four for all datasets except LAR and UAV, which contain three strips -- two parallel and one crossing)
\end{itemize}
 \rev{For each configuration, the GH and DN estimates are compared against (i) the reference values and, where available, (ii) the plane-based on-the-job calibration over the same flight lines. The DN estimation additionally yields the boresight internal correlations to other estimated quantities used to assess parameter observability per IMU quality.}

\medskip\noindent \rev{The following three scenarios characterize the operational robustness of the lightweight GH path, as it is the formulation proposed for routine operator use.}

\medskip\noindent\textbf{Absorption of a constant attitude offset.}
This scenario evaluates the capability of the GH formulation to compensate for a boresight together with a constant orientation bias, such as that caused by imperfect attitude initialization. An artificial angular offset
$\begin{bmatrix}
    \delta_r, \delta_p, \delta_y
\end{bmatrix}$ 
is introduced in the smoothed trajectory prior to georeferencing and correspondence extraction, yielding the biased body-to-map rotation\\
$ \bm R^m_{b^*}(t) = \bm R^m_b(t)\,\bm R^b_{b^*}(\delta_r,\delta_p,\delta_y). $

\medskip\noindent The processing chain described in Sec.~\ref{sec:processing} is then repeated using the biased trajectory. Since a constant IMU orientation bias is mathematically equivalent to an additional constant rotation between the sensor and mapping frames, this experiment evaluates whether such systematic orientation offsets can be absorbed practically by the ``on-the-job'' estimated boresight parameters within the minimum or ``sufficient'' configuration. Two biased datasets are generated with magnitudes representative of realistic residual orientation errors from a tactical-grade ($\sim0.2^\circ$) to a navigation-grade ($\sim0.01^\circ$) IMU. 
\begin{itemize}[noitemsep,topsep=0pt]
    \item Biased LAR: $\begin{bmatrix} \delta_r, \delta_p, \delta_y \end{bmatrix} = \begin{bmatrix} 0.05,\,0.01,\,
0.20 \end{bmatrix} ^\circ $
    \item Biased HAR: $\begin{bmatrix} \delta_r,\delta_p, \delta_y \end{bmatrix} = \begin{bmatrix} 0.20,\,0.05,\,0.01 \end{bmatrix} ^\circ $
\end{itemize}

\begin{figure}[h]
     \centering
     \includegraphics[width=0.8\linewidth]{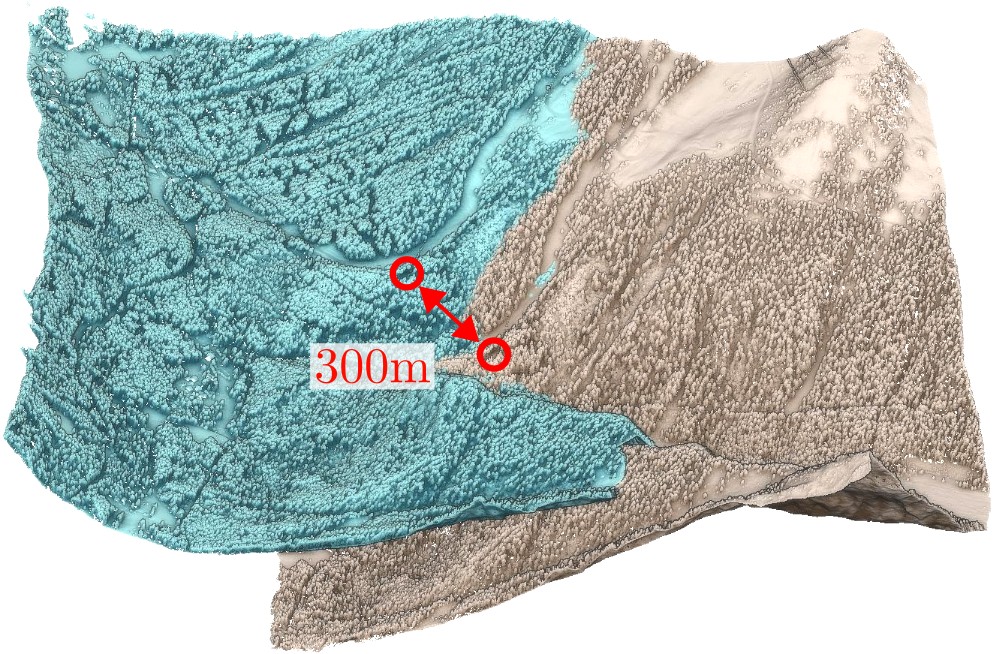}
     \caption{Sample HAR point clouds georeferenced with incorrect boresight of 5$^\circ$ per axis. An example correspondence with points spaced by 300\,m is depicted in red.}
     \label{fig:wrongbor_area}
 \end{figure}

\medskip\noindent\textbf{Sensitivity to initial boresight approximation.}
The robustness of the GH adjustment, and of the upstream correspondence extraction, to large initial boresight errors is investigated by georeferencing the LAR and HAR point clouds using deliberately incorrect boresight angles of up to $5^\circ$ per axis, approximately one order of magnitude larger than typical calibration values. Given the respective flying heights, this misalignment translates into ground georeferencing errors of approximately 40-50\,m for LAR and 250-450\,m for HAR (Fig.~\ref{fig:wrongbor_area}). This experiment evaluates (i) the stability of the correspondence extraction under strong geometrical distortion of the point cloud, (ii) the convergence behavior of the Gauss-Helmert adjustment, and (iii) the practical limits of recoverable boresight.

\medskip\noindent\textbf{Sensitivity to correspondence distribution.}
Finally, the influence of correspondence ``density'' on the stability of the GH boresight estimation is analyzed by temporally downsampling the set of point-to-point correspondences. The correspondence frequency is progressively reduced per flight from 30 correspondences per second (30~Hz, corresponding to an average distance between correspondences of $\sim20$ m) to one every 20 seconds (0.05~Hz, i.e. an average distance of $\sim 800$ m between correspondences). Practically, for each overlapping section, the emission times of all matched points originating from one of the two flight lines are sorted and downsampled to the desired frequency.

This operation reduces the density of geometric constraints while preserving their overall spatial distribution. It therefore serves as a proxy for scenarios with limited surface structure or sparse availability of detected features. Using 5-10 s separation between correspondences reduces also the influence of residual time correlation within the trajectory (GH). The procedure is applied to the HAR scenario with four flight lines and the resulting subsets are introduced into the adjustment model. The stability of estimated parameters and their variances are then analyzed.

\section{Results and analysis}\label{sec:res}
\subsection{On-the-job calibration}
\medskip\noindent \rev{The recovered boresight angles are evaluated by the quality of inertial-sensors, from navigation-grade through tactical-grade to UAV-grade IMUs, each under two flight configurations: a reduced overlap of two strips and a sufficient configuration of four parallel or three (two parallel + one crossing) strips. For every configuration we compare the proposed Gauss-Helmert (GH) and Dynamic Network (DN) estimates using the same correspondences against the reference and, in most cases, also the plane-based (GH-like, trajectory fixed) estimate. Complementary statistics of the lightweight (GH) approach are presented in Tab.~\ref{tab:stats_nominal} for all configurations.}

\subsubsection{Navigation-grade (LAR, HAR)}

\begin{figure}[h!]
    \centering
    \includegraphics[width=7.5cm]{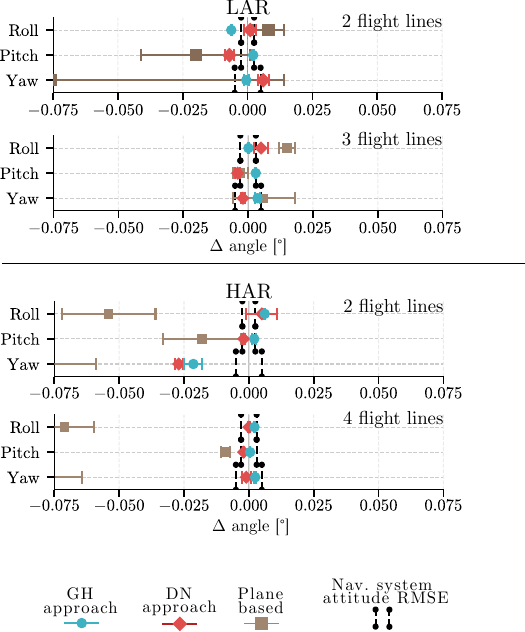}
    \caption{Boresight difference to reference and posterior $3\sigma$ intervals for navigation grade datasets LAR and HAR.}
    \label{fig:nav_diff}
\end{figure}

\begin{figure}[h!]
    \centering
    \includegraphics[width=7.5cm]{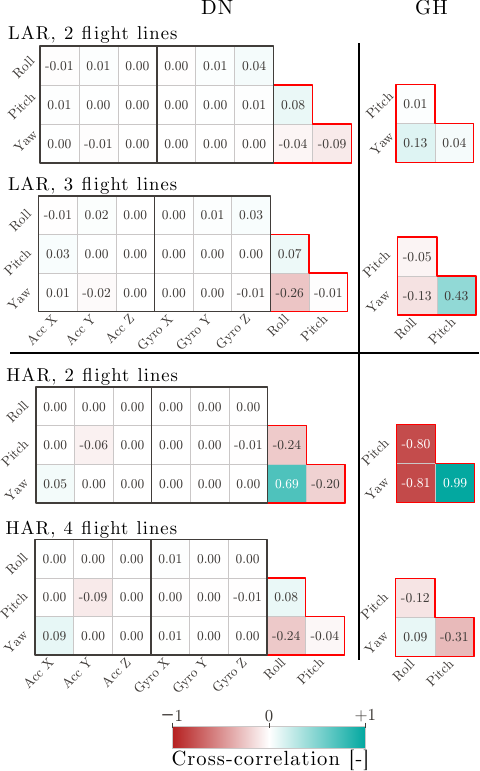}
    \caption{Navigation grade datasets. Left: cross-correlations in DN between lidar boresight angles, accelerometer biases and gyro biases. Right: GH cross-correlation between lidar boresight angles.}
    \label{fig:nav_corr}
\end{figure}

\medskip\noindent\textbf{LAR.}  \rev{For the LAR dataset, the recovered boresight angles via GH closely match the DN estimates, within the uncertainty of the trajectory attitude, in both configurations (Fig.~\ref{fig:nav_diff}, top). The only notable GH difference occurs for roll when only two flight lines are used, with a deviation of $0.004^\circ$. The normal equations are well conditioned (Tab.~\ref{tab:stats_nominal}-2nd column) and the boresight angles remain mutually decorrelated (Fig.~\ref{fig:nav_corr}-right column, 2nd plot), with one exception: a higher pitch–yaw correlation (43\%) appears with the cross strip in GH, though without affecting the recovered angles that remain within the navigation confidence ($3\sigma$). The GH posterior variance factor stays close to unity ($\sigma_0 = 0.3$-$0.5$, Tab.~\ref{tab:stats_nominal} line 3-4), indicating that the stochastic model is consistent with the observed residuals. The joint DN estimation further shows the boresight angles being well separated from the accelerometer and gyroscope biases (Fig.~\ref{fig:nav_corr}-left column), with all boresight-to-bias correlations below $5\%$ in both configurations and low boresight-internal correlations throughout (Fig.~\ref{fig:nav_corr}, top). In contrast, the plane-based calibration deviates largely from the reference at minimal geometry, particularly in yaw ($0.076^\circ$ with two flight lines), reflecting the weaker yaw observability of plane constraints under limited flight geometry despite featuring some urban scenery.}

\medskip\noindent\textbf{HAR.}  \rev{The mountainous, natural-terrain HAR dataset investigates a similar navigation grade system under significantly different scene structure: absence of man-made structures and considerably higher AGL. The estimated angles again agree between GH and DN solutions (Fig.~\ref{fig:nav_diff}, bottom), with roll and pitch recovered within the navigation uncertainty in both configurations. Yaw, however, exposes the geometric requirements: in GH, with two parallel strips the system is weakly conditioned ($\kappa = 926$, Tab.~\ref{tab:stats_nominal} line 7), the boresight angles are strongly correlated ($80-100\%$), and the deviation from the reference yaw angle is too high ($0.021^\circ$) for the stated IMU precision. Introducing two additional strips reduces this divergence within the nominal navigation $1\sigma$ precision ($0.002^\circ$), drastically improves conditioning ($\kappa = 40$) and restores sufficient decorrelation (30\% between pitch and yaw). The same degeneracy is independently visible in the DN solution when using only two strips: boresight-internal correlation is $~70\%$ between roll and yaw. This is resolved in the four-strip configuration (Fig.~\ref{fig:nav_corr}, bottom), confirming that the limitation is geometric rather than specific to either estimator. In DN, the boresight angles nonetheless remain well separated from the inertial biases in both configurations, with boresight-to-bias correlations below $10\%$. Finally, due to the absence of identifiable planar surfaces of a large size, the plane-based method fails to retrieve reliable estimates for any of the three angles in both configurations.}

\subsubsection{Tactical-grade systems (LEG, HAL)}

\begin{figure}[h!]
    \centering
    \includegraphics[width=7.5cm]{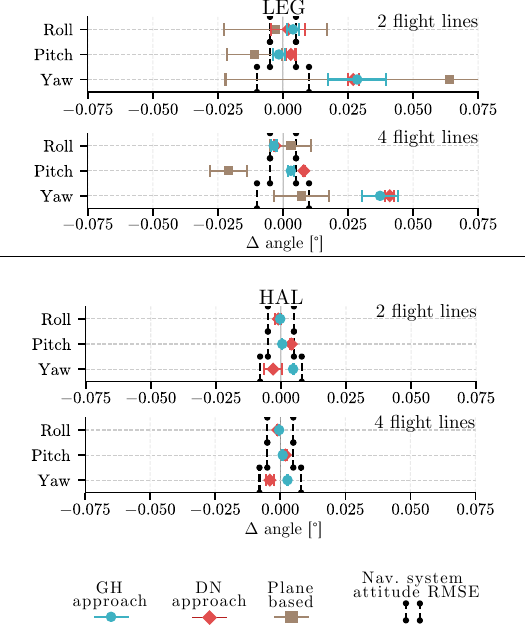}
    \caption{Boresight difference to reference and posterior $3\sigma$ intervals for tactical grade datasets LEG and HAL.}
    \label{fig:tac_diff}
\end{figure}

\begin{figure}[h!]
    \centering
    \includegraphics[width=7.5cm]{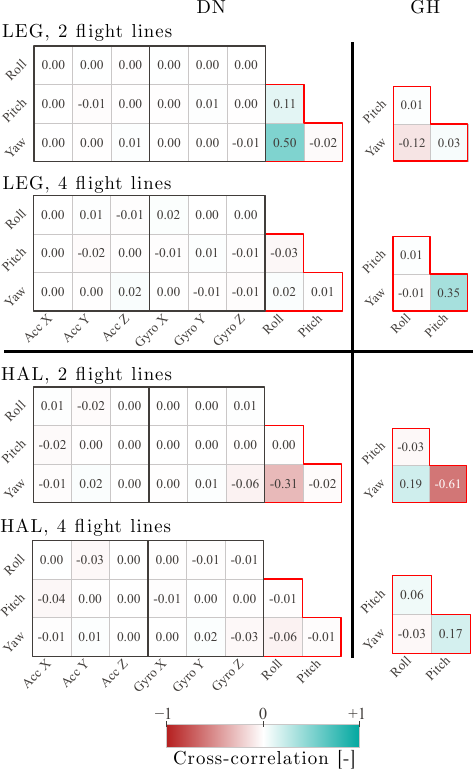}
    \caption{Left: cross-correlations in DN between lidar boresight angles, accelerometer biases and gyro biases. Right: GH cross-correlation between lidar boresight angles, tactical grade datasets.}
    \label{fig:tac_corr}
\end{figure}

\medskip\noindent\textbf{LEG.}  \rev{For the legacy LEG dataset, acquired at low altitude over a built-up area, the GH and DN solutions agree on all estimated boresight angles. Roll and pitch are recovered at a level comparable to the attitude uncertainty in both configurations. The yaw component exhibits a larger discrepancy of approximately $0.03^\circ$ that does not diminish with additional strips (Fig.~\ref{fig:tac_diff}, top). This is attributed to the limited yaw observability at the relatively low flight altitude for a legacy lidar having $7-20\times$ larger beam divergence than the modern sensors, possibly also affecting the reference value itself. The two-strip configuration benefits from a cross strip flown at a different altitude, yielding a relatively well-conditioned system ($\kappa = 26$) despite the reduced overlap (Tab.~\ref{tab:stats_nominal}, rows 1-2). In two-strip configuration, the boresight-internal correlation between roll and yaw remains at $50\%$ in the Dynamic Network; this correlation is negligible ($<3\%$) for all angles when two cross strips are introduced (Fig.~\ref{fig:tac_corr}, top). The boresight angles are also well decorrelated ($<2\%$) from the inertial biases. The GH posterior variance factor stays close to unity ($\sigma_0 = 0.8$-$1.0$). Finally, the plane-based ``on-the-job'' calibration provides a less accurate solution at minimal geometry, with a yaw error of $0.065^\circ$ when two flight lines are used, while remaining biased in pitch ($\sim0.02^\circ$) with 4 flight-lines.}

\medskip\noindent\textbf{HAL.}  \rev{For the HAL dataset, all boresight components are recovered within the navigation uncertainty relative to the reference, with the GH and DN solutions in agreement (Fig.~\ref{fig:tac_diff}, bottom). The GH adjustment converges in a single iteration and remains well conditioned ($\kappa < 25$). The boresight-internal correlations decrease with additional strips, from $60\%$ to $<20\%$ for pitch-yaw in GH and from $30\%$ to $<10\%$ in roll-yaw in DN (Fig.~\ref{fig:tac_corr}, bottom). The boresight angles are well separated from the inertial biases in both configurations, with all boresight-to-bias correlations at or below $6\%$. The posterior variance factor remains close to unity ($\sigma_0 = 0.4$-$0.5$, Tab.~\ref{tab:stats_nominal} line 9-10).}

\subsubsection{UAV-grade system (UAV)}\label{sec:res_uav}

\medskip\noindent  \rev{Because the UAV configuration reuses the LAR lidar, GNSS and flight geometry, replacing only the navigation-grade AIRINS with a UAV-grade APX-15, any difference in boresight observability is attributable to the change of inertial quality alone. The APX-15 stated attitude RMSE is roughly an order of magnitude larger than that of AIRINS and, unlike the navigation-grade IMU, contains significant time-correlated components. The effects of imperfect initialization and time-correlated sensor biases are therefore more pronounced in the attitude obtained by an optimal smoother, which is why the assumption of having ``sufficiently precise'' trajectory for GH boresight recovery may no longer be valid.}

\begin{figure}[h!]
    \centering
    \includegraphics[width=7.5cm]{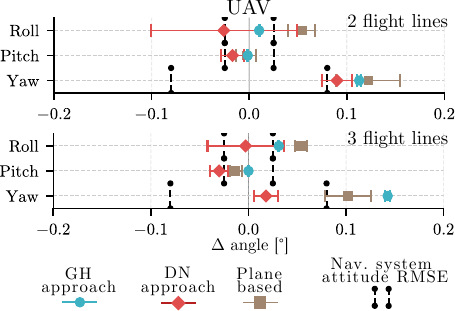}
    \caption{Boresight difference to reference and posterior $3\sigma$ intervals for the UAV grade dataset.}
    \label{fig:uav_diff}
\end{figure}

\medskip\noindent \rev{The GH adjustment converges in a single iteration with conditioning and posterior variance factor comparable to LAR (Tab.~\ref{tab:stats_nominal}). While pitch is indistinguishable from the reference in both configurations, roll is within $1\sigma$ attitude uncertainty with two flight lines ($0.01^\circ$) and within $3\sigma$ with three flight lines ($0.03^\circ$). Yaw, however, and in contrast to the navigation-grade IMU processed over the identical geometry, deviates well beyond the attitude specifications in both configurations ($0.11^\circ$ and $0.14^\circ$). Moreover, the reported posterior $3\sigma$ intervals (e.g. $\pm0.002^\circ$ for yaw) are much smaller than the actual deviation from the reference, indicating that the assumption of temporally uncorrelated trajectory noise underestimates the true parameter uncertainty in this regime. Nevertheless, the point-to-point residuals after adjustment ($0.07-0.14\,m$) remain comparable to the other datasets. In other words, despite the boresight not being recovered correctly, its application to point-cloud georeferencing (with the unchanged INS/GNSS trajectory) yields a geometrically consistent point cloud, suggesting that a significant mean-attitude offset in the trajectory was absorbed by boresight estimation via the GH-approach.} 

\begin{figure}[h!]
    \centering
    \includegraphics[width=7.5cm]{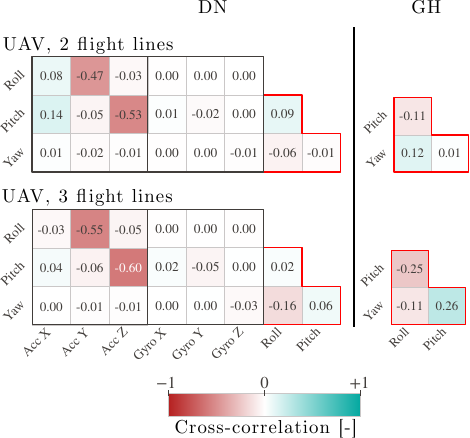}
    \caption{Left: cross-correlations in DN between lidar boresight angles, accelerometer biases and gyro biases. Right: GH cross-correlation between lidar boresight angles, UAV dataset.}
    \label{fig:uav_corr}
\end{figure}

\medskip\noindent \rev{The Dynamic Network solution, which estimates the boresight jointly with the trajectory and the time-correlated inertial errors from raw data, recovers all three angles within the navigation uncertainty when three strips are used, with pitch being marginally above the attitude RMSE (Fig.~\ref{fig:uav_diff}). The DN posterior $3\,\sigma$ intervals scale consistently with the trajectory quality (an order of magnitude larger than those reported by GH), reflecting a stochastic model in better agreement with the actual error sources.} 

\medskip\noindent \rev{The plane-based calibration follows the same pattern as GH, with pitch close to the reference, yet roll and yaw biased and overly optimistic posterior $\sigma$. This similarity is expected, since both the GH and plane-based formulations treat the smoothed trajectory as fixed and therefore absorb the same time-correlated attitude errors into the estimated boresight, whereas the Dynamic Network estimates the trajectory jointly with boresight.}

\medskip\noindent \rev{The covariance analysis makes the mechanism explicit (Fig.~\ref{fig:uav_corr}). In contrast to all higher-grade datasets, where the boresight angles are well separated from the inertial biases ($<15\%$), the UAV configuration exhibits a modest $\sim50$-$60\%$ coupling with some accelerometer biases, while the correlations with gyroscope biases remain insignificant. Such coupling does not diminish when a third strip is introduced, confirming that the limited separability is set more by the inertial error characteristics (although those are not detailed by the manufacturer) rather than by the flight geometry. 
Nevertheless, the Dynamic Network, modeling these biases explicitly, partially disentangles them from the boresight, whereas the GH formulation can only absorb their mean effect into the estimated angles. 
This effect is further examined in Sec.~\ref{sec:disc}.}

\begin{table}[h]
\small
\centering
\begin{tabular}{
l
S[table-format=3.0, round-precision=0]
S[table-format=1.1, round-precision=1]
S[table-format=2.1, round-precision=1]
S[table-format=1.2]
S[table-format=1.2]
}
\toprule
& & & \multicolumn{3}{c}{Residuals [m]}  \\
\cmidrule(lr){4-5} 
Dataset & {$\kappa(J)$} & {$\sigma_0$} & {Init.} & {Final} \\
\midrule
LEG 2 & 26  & 0.9 & 0.87  & 0.15 \\
LEG 4 & 48  & 1.1 & 1.03  & 0.14 \\
\arrayrulecolor{black!30}\midrule\arrayrulecolor{black}
LAR 2 & 13  & 0.5 & 1.28  & 0.04 \\
LAR 3 & 21  & 0.4 & 1.27  & 0.06 \\
\arrayrulecolor{black!30}\midrule\arrayrulecolor{black}
UAV 2 & 15  & 0.3 & 1.25  & 0.07 \\
UAV 3 & 28  & 0.5 & 1.25  & 0.14 \\
\arrayrulecolor{black!30}\midrule\arrayrulecolor{black}
HAR 2 & 926 & 0.9 & 6.80  & 0.15 \\
HAR 4 & 40  & 0.8 & 12.19 & 0.18 \\
\arrayrulecolor{black!30}\midrule\arrayrulecolor{black}
HAL 2 & 24  & 0.4 & 5.07  & 0.14 \\
HAL 4 & 15  & 0.6 & 5.99  & 0.20 \\
\bottomrule
\end{tabular}
\caption{Additional adjustment metric of the GH on all five system configurations: condition number ($\lambda_{max}/\lambda_{min}$), posterior $\sigma_0$, initial and final point-to-point residuals.}
\label{tab:stats_nominal}
\end{table}
\subsection{Operational employment of the GH adjustment}\label{sec:gh_robustness}

\medskip\noindent \rev{The preceding section established the accuracy and observability of estimated boresight in both formulations across IMU grades. The following experiments characterize the operational aptitude of the lightweight GH formulation specifically, as it is the proposed formulation when only a post-processed trajectory is available. Three properties are examined: its ability to absorb a constant attitude offset present in the trajectory, its convergence under large boresight angles, and its stability as the set of established point-to-point correspondences is largely reduced. The low sensitivity of the DN-trajectory to sparse and noisy point-to-point observations is not examined here as it is reported in \citet{brun2022, Brun2025gen}.}

\subsubsection{Absorption of a constant attitude offset}
Whereas the UAV configuration (Sec.~\ref{sec:res_uav}) showed the GH formulation absorbing only the mean of a \emph{time-varying} attitude error, this experiment isolates its behavior under a genuinely \emph{constant} offset, synthetically injected into the smoothed trajectory prior to georeferencing and correspondence extraction. This validates practically the scene-agnostic properties of correspondences within different environments under large and varying effects on ground (attitude versus distance to terrain), while the difference in the GH-estimated boresight is compared against the sum of the true boresight and the injected offset. The processing chain of Sec.~\ref{sec:processing} is otherwise unchanged.

\noindent For both the LAR and HAR datasets, the combined offset is recovered for all three angles regardless of its magnitude (from $0.01^\circ$ to $0.2^\circ$), with the difference between the expected and estimated boresight remaining below $0.001^\circ$ in all cases (Tab.~\ref{tab:bias_bor}). The establishment of correspondences was not affected, and no significant change is observed in the conditioning of the adjustment ($\kappa < 40$), the number of iterations (1), or the cross-correlations between parameters ($<15\%$).  

\noindent This confirms that the GH formulation absorbs a constant rotational offset, whether of boresight or of residual constant-attitude origin, without degradation, and validates the numerical robustness of the correspondence extraction and adjustment under such an offset. It does not, however, address time-varying trajectory errors: the contrast with the UAV case (Sec.~\ref{sec:res_uav}) is precisely that there the same formulation could absorb only the mean component of a time-correlated error, while the DN path was required to recover the boresight.
\vspace{2mm}

\noindent\begin{minipage}{\linewidth}
\centering
\resizebox{\columnwidth}{!}{
\begin{tabular}{cccccc} 
\toprule
ID & \begin{tabular}[c]{@{}c@{}}IMU bias +\\bor. [$^\circ$]\end{tabular} & \begin{tabular}[c]{@{}c@{}}Estimated\\bor. [$^\circ$]\end{tabular} & \begin{tabular}[c]{@{}c@{}}|Diff.|\\{[}$^\circ$]\end{tabular} & \begin{tabular}[c]{@{}c@{}}Init.\\res. [m]\end{tabular} & \begin{tabular}[c]{@{}c@{}}Final\\res. [m]\end{tabular} \\ 
\midrule
\multirow{3}{*}{LAR} & -0.162 & -0.163 & 0.001 & \multirow{3}{*}{1.05} & \multirow{3}{*}{0.06} \\
 & 0.113 & 0.114 & 0.001 &  &  \\
 & 0.393 & 0.393 & \textless 0.001 &  &  \\ 
\hline
\multicolumn{1}{l}{\multirow{3}{*}{HAR}} & 0.408 & 0.408 & \textless 0.001 & \multirow{3}{*}{22.67} & \multirow{3}{*}{0.17} \\
\multicolumn{1}{l}{} & 0.075 & 0.075 & \textless 0.001 &  &  \\
\multicolumn{1}{l}{} & 0.363 & 0.363 & \textless 0.001 &  &  \\
\bottomrule
\end{tabular}
}
\captionof{table}{Adjustment statistics for recovering boresight together with an injected constant attitude bias.}\label{tab:bias_bor}
\end{minipage}

\subsubsection{Sensitivity to initial boresight approximation}
This experiment evaluates the sensitivity of the GH adjustment, and of the upstream correspondence extraction, to the initial mounting matrix $\bm T_s^{b*}$. We consider deviations up to $\pm5^\circ$ per axis, an order of magnitude larger than typical ALS boresight angles, inducing point displacements from 30\,m to 60\,m for the LAR flight and from 250\,m to 450\,m for the HAR flight. The objective is to demonstrate that both the detection stage \citep{Brun2025gen} and the simplified adjustment can cope with such a coarse approximation across different ALS scenes.% and projection effects causing spatial variations.

For the LAR dataset (Tab.~\ref{tab:wrong_bor}), the initial point-to-point residuals are reduced from 44\,m to 0.17\,m after a single least-squares iteration, with no significant change in conditioning or parameter correlations relative to the nominal initialization; the recovered angles remain within the navigation attitude uncertainty. Due to the higher flight altitude, the HAR experiment exhibits much larger initial residuals when the point cloud is georeferenced with the incorrect $5^\circ$ boresight, approximately 350\,m on average. Nevertheless the correspondences are correctly identified, the adjustment converges in a single iteration with conditioning and cross-correlations comparable to the nominal case, and the recovered solution remains within the navigation uncertainty for pitch and yaw, with a slightly larger deviation of $0.004^\circ$ for roll.
\vspace{2mm}

\noindent\begin{minipage}{\linewidth}
\centering
\small
\resizebox{\columnwidth}{!}{
\begin{tabular}{ccccccc} 
\toprule
\multicolumn{1}{c}{ID} & \multicolumn{1}{c}{\begin{tabular}[c]{@{}c@{}}Apriori\\bor. [$^\circ$]\end{tabular}} & \multicolumn{1}{c}{\begin{tabular}[c]{@{}c@{}}Estimated\\bor. [$^\circ$]\end{tabular}} & \multicolumn{1}{c}{\begin{tabular}[c]{@{}c@{}}Diff.\\{[}$^\circ$]\end{tabular}} & \multicolumn{1}{c}{\begin{tabular}[c]{@{}c@{}}$\sigma_{rpy}$\\{[}$^\circ$]\end{tabular}} & \multicolumn{1}{c}{\begin{tabular}[c]{@{}c@{}}Initial\\res. [m]\end{tabular}} & \multicolumn{1}{c}{\begin{tabular}[c]{@{}c@{}}Final\\res. [m]\end{tabular}} \\ 
\midrule
\multirow{3}{*}{LAR} & 5 & -0.213 & -0.001 & 0.001 & \multirow{3}{*}{44.15} & \multirow{3}{*}{0.17} \\
 & 5 & 0.106 & 0.003 & 0.001 &  &  \\
 & 5 & 0.193 & -0.001 & 0.003 &  &  \\ 
\midrule
\multirow{3}{*}{HAR} & 5 & 0.210 & 0.004 & 0.000 & \multirow{3}{*}{348.28} & \multirow{3}{*}{0.42} \\
 & 5 & 0.024 & -0.001 & 0.000 &  &  \\
 & 5 & 0.352 & 0.000 & 0.001 &  &  \\
\bottomrule
\end{tabular}
}
\captionof{table}{Adjustment statistics for the GH calibration with an incorrect initial boresight of $5^\circ$ per axis.}\label{tab:wrong_bor}
\end{minipage}

\subsubsection{Sensitivity to correspondence distribution}\label{sec:down_freq}
\begin{figure}[h]
    \centering
    \includegraphics[width=\linewidth]{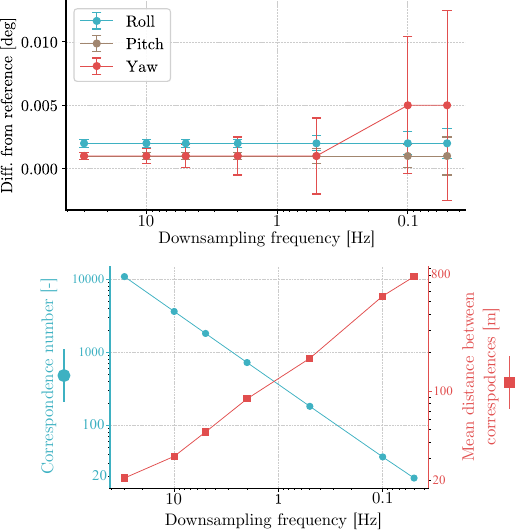}
    \caption{Top: estimated boresight and posterior uncertainty versus correspondence downsampling frequency (HAR, four strips). Bottom: number of correspondences and corresponding average distance between them versus downsampling frequency.}
    \label{fig:downsample_freq}
\end{figure}
Finally, we examine how many correspondences the GH adjustment requires, by temporally down-sampling the correspondence set from $\sim$30~Hz to $\sim$0.05~Hz. This reduces the ground density of geometric constraints while preserving their spatial distribution, serving as a proxy for scenes with limited surface structure or sparse detectable features.

Top of Fig.~\ref{fig:downsample_freq} shows the evolution of the estimated boresight and its posterior uncertainty as the correspondence rate decreases from 30~Hz ($\sim$10\,800 correspondences) to one every 20\,s of flight (20 correspondences). The bottom panel translates this rate into the resulting on-map correspondence density: the mean three-dimensional spacing between neighboring correspondences grows from $\sim$20\,m at full rate to $\sim$800\,m at the sparsest setting, a roughly 40$\times$ increase in spacing. Roll and pitch remain essentially unaffected across this entire range. Yaw begins to deviate only at 0.1~Hz ($\sim$40 correspondences, $\sim$550\,m mean spacing), and by $0.005^\circ$ only, remaining within the attitude uncertainty for this angle;
its posterior $3\sigma$ interval widens from $\pm0.001^\circ$ at full rate to $\pm0.009^\circ$ at the sparsest configurations.
\section{Discussion}\label{sec:disc}

\rev{\textbf{Overall assessment.} Across the investigated navigation and tactical-grade systems, the lightweight Gauss-Helmert and rigorous Dynamic Network formulations yield boresight estimates that agree within the attitude uncertainty, and the joint estimation confirms that the boresight angles are decorrelated from the inertial biases (correlations below $15\%$). The adequacy of the adopted stochastic model, including the empirical weighting $\sigma_{p2p}=\tfrac{1}{2}\mathrm{GSD}$ inherited from \cite{Brun2025gen}, is further supported by the GH posterior variance factors remaining close to unity across all configurations (Tab.~\ref{tab:stats_nominal}), indicating consistency between the assumed and observed error levels.
This agreement is not incidental: the GH adjustment is a narrowed adjustment setup in which the smoothed INS/GNSS trajectory is assumed sufficiently precise to serve as fixed observations, with residual navigation errors absorbed as stochastic noise, whereas the DN makes no such assumption and estimates the boresight jointly with the trajectory and the inertial errors from raw data. Their consistency therefore provides an empirical validation of the trajectory-sufficiency assumption: when the optimal smoother produces a trajectory of adequate quality, narrowing the joint estimation (DN) onto the boresight alone (GH) discards nothing and returns the correct boresight while remaining considerably simpler and convenient to operate. The UAV-grade configuration marks the boundary of this regime: although the UAV configuration shares the platform, lidar and flight geometry of its navigation-grade counterpart, differing only in the inertial sensor, its boresight couples strongly to the accelerometer errors ($50\%$-$60\%$). As these biases vary in time, this coupling cannot be mitigated by additional strips, and the two formulations diverge in yaw. Where the trajectory cannot be trusted as a fixed observation, only the rigorous path of DN that models the time-varying effects of inertial errors (observability of which is improved through lidar correspondences), explicitly recovers the boresight, while the lightweight formulation can only absorb their mean effect. As shown, this may still yield a geometrically consistent point cloud over the duration of the mission. This boundary delineates the operating envelope developed in the remainder of this section.}

\medskip\noindent
\rev{\textbf{Yaw angle observability.} In all configurations the yaw boresight is the hardest of the three angles to observe, for two complementary reasons. The first is purely geometric. A roll or pitch error rotates the laser vector primarily along its dominant downward component, whereas a yaw error displaces it only within the horizontal plane; since the across-track field of view is typically $30$-$35^\circ$ per side, the downward component of the laser vector is usually two to three times larger than the across-track one. Consequently, a yaw error of a given magnitude propagates into a smaller, less observable point displacement than the same error in roll or pitch. This weakness is relieved only by geometric diversity between strips: with parallel lines the across-track displacements are nearly collinear and the yaw component remains poorly constrained, whereas a crossing line, or an altitude-separated line, or a fourth parallel line introduces the diversity needed to separate it. This is the mechanism behind the conditioning and yaw behavior observed with navigation and tactical IMU grades, where two parallel strips leave yaw weakly observable and two additional strips yield clear and sufficient improvements.}

\noindent 
\rev{The second reason is inertial error characteristics, and is the one that distinguishes the UAV grade experiment. The heading error of an INS/GNSS system is only observable through the comparison of GNSS-observed against accelerometer-sensed velocity changes, which is weakly observable on an IMU flown at near-constant velocity and heading \citep{Hong2025ObsAnal} and therefore even lower on platform-stabilized systems. At navigation and tactical grades, provided the trajectory attains its nominal post-processed quality (sufficient satellite coverage, correct initialization and alignment) the residual heading error is small enough that the boresight yaw remains separable from the accelerometer biases. This is not true at UAV grade, where the observed coupling is the direct consequence of this limit. 
The coupling is consequently set by the inertial error characteristics and the flight dynamics, not only by the strip geometry and additional strips, which no longer sufficiently improve observability in this scenario. In other words, geometric diversity resolves the limited geometric yaw observability but cannot compensate for weak inertial observability within the filtered/smoothed trajectory.
}

\medskip\noindent
\rev{\textbf{Absorption of attitude bias.} A constant boresight misalignment and a constant residual attitude error have the same effect on the georeferenced point cloud, both manifesting as a fixed rotation between the sensor and mapping frames. The GH formulation should therefore be understood as estimating a combined constant rotational offset, rather than separately recovering two distinct quantities. The synthetic-offset experiment confirms this directly: an injected constant orientation bias is absorbed into the estimated boresight within $0.001^\circ$ regardless of its magnitude, without degradation of conditioning or convergence, since from the point cloud's perspective the injected bias and the true boresight are one and the same rotation. This absorption is however operationally useful, yielding a geometrically consistent point cloud even when a constant attitude bias is present, but without separation of the two effects.}

\rev{\noindent The distinction between both approaches materializes when the attitude error is not constant, as illustrated in the UAV-grade case. There, the time-correlated heading error cannot be absorbed as a single fixed offset, and the GH boresight captures only its mean component, leaving the recovered yaw-angle biased beyond the navigation uncertainty; the resulting point cloud remains usable, though its residuals are roughly twice those of its navigation-grade counterpart over identical geometry, reflecting the time-varying component that the GH fixed-trajectory adjustment cannot absorb. Recovering the boresight in this regime requires explicitly modeling the inertial error sources and estimating them jointly with the boresight, as in a Dynamic Network. The two formulations thus occupy distinct regimes of separability: where the residual attitude error is constant, GH absorbs it into an effective boresight without needing to separate the two; where it is time-varying, only the rigorous DN path separates the boresight from the inertial error, while GH retains its value as a means of absorbing the mean effect for georeferencing.}

\medskip\noindent
\rev{\textbf{Operational robustness.} Beyond the formerly discussed accuracy limits, the GH adjustment is operationally forgiving in two important aspects that matter for routine use. First, it requires no accurate prior knowledge of the boresight: even when the point clouds are initially georeferenced with deliberately incorrect angles of $5^\circ$ per axis, producing misalignments of several hundred meters on the ground, the correspondences are still established and the adjustment recovers the correct parameters in a single iteration, with conditioning and correlations comparable to the nominal case. Second, the estimation remains stable under a severe reduction of the available constraints: roll and pitch are essentially unaffected when the correspondence density is reduced by nearly three orders of magnitude, and yaw exhibits only a moderate increase in uncertainty. This indicates that the method remains effective in scenes offering very limited geometric structure, provided the available correspondences are reasonably distributed across the overlap. As a rule of thumb only tens, not hundreds or thousands, are required.}

\medskip\noindent\rev{
\textbf{Operating envelope.} The preceding results delineate the conditions under which boresight can be \textit{calibrated on the job}, without dedicated patterns. Three requirements emerge. First, and for GH-approach, the INS/GNSS trajectory must be of post-processed quality obtained under nominal conditions (continuous satellite coverage, correct initialization and alignment) so that the residual attitude error is effectively at the random noise level over the acquisition. Second, for both approaches, the strip configuration must provide sufficient geometric diversity to render boresight in yaw observable: in our experience four parallel strips, or three including a crossing or altitude-separated line. Third, the scene must return reliable lidar echoes and offer enough geometric structure for correspondence extraction, which excludes water, wet or otherwise lidar reflection absorbing terrain but, as shown across urban, mixed and natural sites, imposes no requirement on the form of observed surfaces. Within this envelope, the two formulations are equivalent and the lightweight GH adjustment is the appropriate tool, returning the correct result rapidly and conveniently. The envelope's boundary is reached when the trajectory cannot be trusted as a fixed observation affected only by random noise: where its quality cannot be guaranteed, or where time-varying attitude errors are suspected; as at UAV grade where geometric diversity cannot compensate for the weak inertial observability. There, boresight estimation within the Dynamic Network from raw inertial data is the rigorous and recommended path. However, the influence of mean attitude error and boresight can still be absorbed by the GH approach, yet without the capacity to separate them.}

\medskip\noindent
\rev{
\textbf{Comparison with plane-based calibration.} Within the described operating envelope, the point-to-point formulation offers two practical advantages over the conventional plane-based approach. It requires only three to four overlapping strips, which may follow a standard parallel surveying pattern rather than a dedicated cross-pattern at several altitudes; and it imposes no prerequisite on the scene beyond sufficient geometric structure for correspondence extraction, removing the dependence on well-defined planar surfaces of varied orientation. The benefit is most visible where planar features are scarce: on the natural mountainous terrain of HAR, the plane-based calibration fails to recover reliable angles in any configuration, whereas the point-to-point adjustment succeeds. It should be noted that the plane-based and GH formulations share the same reliance on a trajectory treated as fixed, and therefore the same vulnerability to time-correlated attitude errors, as both absorb them identically into the estimated boresight; the advantage of the point-to-point formulation lies in lifting the scene prerequisite, not in altering this shared dependence. Recovering the boresight when such dependence fails remains the role of the Dynamic Network.}

\rev{
\section{Conclusion}\label{sec:conclusion}
\noindent We have presented a scene-agnostic approach to ALS boresight self-calibration built on point-to-point correspondences, in two complementary formulations: a lightweight Gauss-Helmert (GH) adjustment that treats the smoothed INS/GNSS trajectory as observations affected by zero mean random noise, and a rigorous Dynamic Network (DN) path that estimates the boresight jointly with the trajectory and the time-correlated inertial errors from raw data. Evaluated across four operational ALS flights spanning five commercial inertial system across navigation, tactical and UAV-grades, the two formulations agree wherever the trajectory is of sufficient quality and diverge precisely where it is not, delineating an operating limit of the former approach when \textit{calibration} changes to \textit{absorption} of a combined effect (mean attitude error plus boresight).
}

\noindent
\rev{The specific contributions of this work are the following. The GH formulation provides a parametric boresight calibration from point-to-point correspondences that removes the dependence on scenes rich in planar surfaces required by the conventional approach \citep{Skaloud2006b}; its benefit is clearest on exclusively natural terrain such as HAR or terrain where presence of forest is important such as LAR, where the plane-based method fails to recover reliable angles and the point-to-point adjustment succeeds. From the observability analysis we derive the minimal flight-line configuration that renders the boresight separable under nominal trajectory quality, namely four parallel strips or three including a crossing or altitude-separated line. The UAV-grade case further shows that this geometric condition is necessary but not sufficient, as flight-line diversity cannot compensate for weak GNSS/INS observability when the trajectory carries significant time-correlated errors. We provide the first quantitative evaluation of boresight estimation within the rigorous Dynamic Network across these inertial grades, including the boresight-to-biases correlation analysis, and we validate the lightweight GH formulation over the same envelope for navigation and tactical grade inertial systems. For the UAV-grade system, we further demonstrate that the GH formulation absorbs the boresight together with the mean of a time-correlated attitude error, still yielding a usable point cloud while, as expected, no longer separating the two effects.}

\noindent \rev{Both formulations are released for use and verification. The point-to-point correspondences, raw inertial, GNSS and lidar observations, approximate and adjusted trajectories and point clouds, together with the configuration files needed to reproduce the Dynamic Network results through a freely accessible service, are made available for three of the flights; the Gauss-Helmert adjustment is released as an open-source implementation together with its  configurations for the described experiments.
}
\newline

\noindent
\begin{tabular}{@{}l p{0.55\linewidth}@{}}
\toprule
Boresight calibration code & \href{https://github.com/ESO-EPFL/libor}{github.com/ESO-EPFL/libor} \\
ALS data & \href{https://addlidar.epfl.ch}{addlidar.epfl.ch} \\
Online DN solver & \href{https://odyn.epfl.ch/}{ODyN.epfl.ch} \\
\bottomrule
\end{tabular}\label{tab:links}
\newline\vspace{1pt}

\noindent \rev{Taken together, these results support the practical conclusion that ALS boresight calibration can be performed within routine mapping operations via the Gauss-Helmert adjustment, without dedicated flight patterns or scene prerequisites, provided that the conditions of the operating envelope are met. Where they are not, e.g. due to degraded trajectory quality or suspected time-varying attitude errors, the Dynamic Network remains the rigorous path, recovering the boresight from raw inertial data while the lightweight formulation can only absorb its mean effect.}

\section*{Acknowledgment}
This contribution was supported by the InnoSuisse project 119.293 IP-ENG. %Part of the flights were provided by Sixense Helimap SA. The personal involvement and contribution of its former director, Julien Vallet are highly appreciated. The authors would like to thank Jesse LaHaye for his help in proofreading the manuscript.

%\bibliography{References.bib}
\bibliography{JanBib.bib}
\clearpage
\onecolumn
\appendix{Derivation of the point-to-point constraint's Jacobians}
\textbf{Derivation of $A$.}  
Matrix $A$ is the partial derivatives of $f$ with respect to the unknown boresight angles $\bm{\theta}=[\alpha,\beta,\gamma]^\top$. For one point-to-point difference with points ${{i,\,j}}$, rearranging the terms in (\ref{eq:f}) gives:

\begin{equation}
f(\boldsymbol\theta, \bm{\ell}_{ij})
= \mathbf{p}_i^m(\boldsymbol\theta) - \mathbf{p}_j^m(\boldsymbol\theta) + \mathbf{w}
= \big(\mathbf{n}_i^m - \mathbf{n}_j^m\big)
   + \big(R_{b,i}^m\mathbf{u}_i - R_{b,j}^m\mathbf{u}_j\big) 
   + \big(R_{b,i}^m - R_{b,j}^m\big)\mathbf{a}^b
   + \big(R_{b,i}^m U_i - R_{b,j}^m U_j\big)\,\boldsymbol\theta + \mathbf{w}.
\end{equation}

and the derivative follows:
\begin{equation}
\underset{3\times3}{A_k} = \frac{\partial f}{\partial \boldsymbol\theta} \;=\; R_{b,i}^m U_i \;-\; R_{b,j}^m U_j.
\end{equation}

\textbf{Derivation of $B$.} Matrix $B$ collects the partial derivatives of $f$ with respect to the observations. 
Introducing $\mathbf{s}_k = \mathbf{u}_k + U_k \bm{\theta} + \mathbf{a}$ to simplify the notation, the model becomes:
$
\boxed{f(\boldsymbol\theta, \bm{\ell}_{ij})
= (\mathbf{n}_i^m - \mathbf{n}_j^m)
\;+\; R_{b,i}^m\,\mathbf{s}_i
\;-\; R_{b,j}^m\,\mathbf{s}_j.
}
\label{eq:factor_direct}
$

From there, the derivative with respect to the different observations follows:
\begin{align}
    \frac{\partial f}{\partial [X,Y,Z]_i}  &= I_3, & \frac{\partial f}{\partial [X,Y,Z]_j}& = -I_3, \\
    \frac{\partial f}{\partial r_i} &=\frac{\partial R_{b,i}^m}{\partial r_i}\,\mathbf{s}_i, &
    \frac{\partial f}{\partial r_j} &=-\frac{\partial R_{b,j}^m}{\partial r_j}\,\mathbf{s}_j, \\
    \frac{\partial f}{\partial p_i} &=\frac{\partial R_{b,i}^m}{\partial p_i}\,\mathbf{s}_i, &
    \frac{\partial f}{\partial p_j} &=-\frac{\partial R_{b,j}^m}{\partial p_j}\,\mathbf{s}_j, \\
    \frac{\partial f}{\partial y_i} &=\frac{\partial R_{b,i}^m}{\partial y_i}\,\mathbf{s}_i,&
    \frac{\partial f}{\partial y_j} &=-\frac{\partial R_{b,j}^m}{\partial y_j}\,\mathbf{s}_j,
\end{align}
And $B_k$ is formed by stacking these contributions into a $3\times 12$ Jacobian
\begin{equation}
    \underset{3\times12}{B_k} = 
    \begin{bmatrix}
        I_3 & \frac{\partial R_{b,i}^m}{\partial r_i}\mathbf{s}_i &
               \frac{\partial R_{b,i}^m}{\partial p_i}\mathbf{s}_i &
               \frac{\partial R_{b,i}^m}{\partial y_i}\mathbf{s}_i &
       -I_3 & -\frac{\partial R_{b,j}^m}{\partial r_j}\mathbf{s}_j &
              -\frac{\partial R_{b,j}^m}{\partial p_j}\mathbf{s}_j &
              -\frac{\partial R_{b,j}^m}{\partial y_j}\mathbf{s}_j
    \end{bmatrix}.
\end{equation}

\end{document}